\documentclass[letterpaper, 10 pt, conference]{ieeeconf}  %

\IEEEoverridecommandlockouts                              %

\usepackage{amsmath} %
\usepackage{amssymb}  %
\usepackage{cite}
\usepackage{algorithm}
\usepackage[noend]{algorithmic}
\usepackage{graphicx}
\usepackage{bm}
\usepackage{eso-pic}
\usepackage{xspace}
\usepackage{tabularx}
\usepackage{array}
\usepackage{multicol}
\usepackage{booktabs}
\usepackage{flushend}
\usepackage{dsfont}
\usepackage{subfig}
\usepackage{url}
\usepackage{hyperref}
\usepackage[capitalize]{cleveref}
\crefname{section}{Sec.}{Secs.}
\Crefname{section}{Section}{Sections}
\Crefname{table}{Table}{Tables}
\crefname{table}{Tab.}{Tabs.}
\crefname{algorithm}{Alg.}{Agls.}
\Crefname{section}{Algorithm}{Algorithms}

\newcommand{\method}{CTG\xspace}

\definecolor{tblue}{HTML}{1F77B4}
\definecolor{tred}{HTML}{FF6961}
\definecolor{tgreen}{HTML}{429E9D}
\definecolor{thighlight}{HTML}{000000}
\newcolumntype{P}{>{\raggedleft\arraybackslash}X}

\definecolor{cred}{HTML}{D62728}
\definecolor{cblue}{HTML}{1F77B4}
\definecolor{cgreen}{HTML}{79AB76}
\definecolor{cgrey}{rgb}{0.6,0.6,0.6}

\definecolor{highlight}{rgb}{0,0,0}

\newcommand{\bc}{\mathbf{c}}
\newcommand{\btau}[1]{\bm{\tau}^{#1}}
\newcommand{\opt}{\mathcal{O}}

\newcommand{\bmu}{\boldsymbol{\mu}}
\newcommand{\bSigma}{\boldsymbol{\Sigma}}

\renewcommand{\bmu}{\bm{\mu}}

\renewcommand{\bSigma}{\bm{\Sigma}}

\title{\LARGE \bf
Guided Conditional Diffusion for Controllable Traffic Simulation
}

\author{Ziyuan Zhong, Davis Rempe, Danfei Xu, Yuxiao Chen, \\Sushant Veer, Tong Che, Baishakhi Ray, and Marco Pavone \vspace{3mm}\\
\thanks{Ziyuan Zhong is with Columbia University, this work was carried out while at NVIDIA,
{\tt\small ziyuan.zhong@columbia.com}. Davis Rempe is with Stanford University, this work was carried out while at NVIDIA, {\tt\small drempe@stanford.edu}. Danfei Xu is with Georgia Tech, and with NVIDIA Research,
{\tt\small danfei@gatech.edu, danfeix@nvidia.com}. Yuxiao Chen, Sushant Veer, Tong Che are with NVIDIA Research, {\tt\small yuxiaoc, sveer, tongc@nvidia.com}. Baishakhi Ray is with Columbia University,
{\tt\small rayb@cs.columbia.edu}. Marco Pavone is with Stanford University, and with NVIDIA Research, {\tt\small pavone@stanford.edu, mpavone@nvidia.com}}%
}

\begin{document}

\maketitle
\thispagestyle{empty}
\pagestyle{empty}

\begin{abstract}
Controllable and realistic traffic simulation is critical for developing and verifying autonomous vehicles.
Typical heuristic-based traffic models offer flexible control to make vehicles follow specific trajectories and traffic rules.
On the other hand, data-driven approaches generate realistic and human-like behaviors, improving transfer from simulated to real-world traffic.
However, to the best of our knowledge, no traffic model offers both controllability and realism.
In this work, we develop a conditional diffusion model for controllable traffic generation (CTG) that allows users to control desired properties of trajectories at test time (e.g., reach a goal or follow a speed limit) while maintaining realism and physical feasibility through enforced dynamics.
The key technical idea is to leverage recent advances from diffusion modeling and differentiable logic to guide generated trajectories to meet rules defined using signal temporal logic (STL).
We further extend guidance to multi-agent settings and enable interaction-based rules like collision avoidance.
CTG is extensively evaluated on the nuScenes dataset for diverse and composite rules, demonstrating improvement over strong baselines in terms of the controllability-realism tradeoff.
\end{abstract}

\section{Introduction}
Simulation is crucial to comprehensively evaluate modern autonomous vehicles (AVs). Due to the difficulty and danger of running large-scale real-world tests~\cite{nummilesneeded},
AV developers rely on extensive testing in simulation to produce reliable systems~\cite{waymoreport}.
To be most useful, simulators must embody both \textit{realism} and \textit{controllability}, especially for the models of traffic.
\textit{Realistic} traffic allows developments made in simulation to faithfully transfer to the real world, while \textit{controllability} enables constructing fine-grained traffic scenarios to analyze specific AV behavior.
Yet, developing realistic traffic models is still an open challenge\cite{Suo_2021_CVPR,bits2022}, and little attention has been devoted to making such models easily controllable.

\begin{figure}[ht]
\centering
\includegraphics[width=0.4\textwidth]{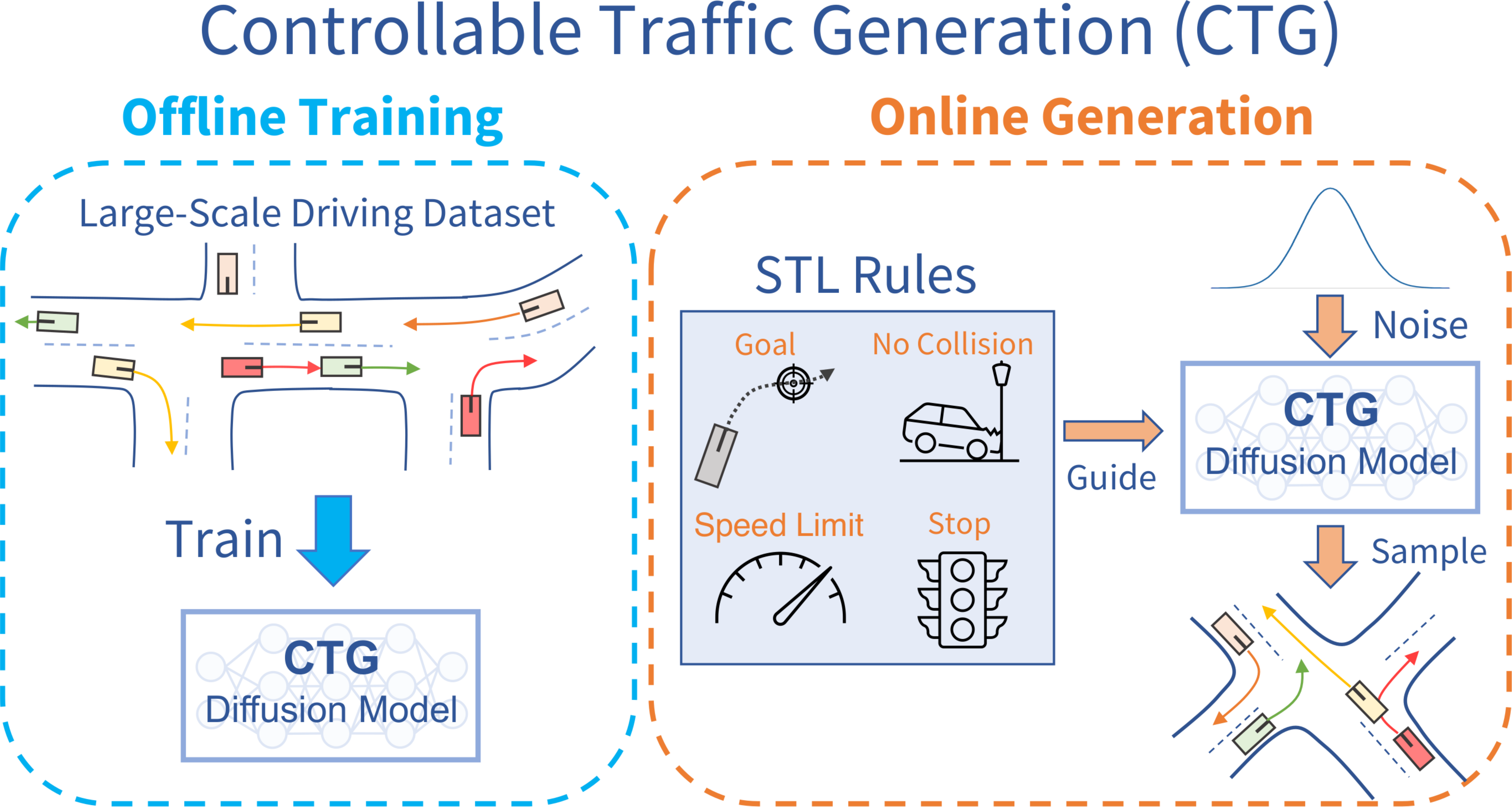}
\caption{\textbf{\small{Overview.}} \small{CTG uses two key stages to enable controllable simulation. (Left) a conditional diffusion model is trained to generate realistic trajectories. (Right) Guided model sampling uses STL rules to meet desired objectives.}}
\label{fig:overview}
\vspace{-8mm}
\end{figure}

Existing driving simulators commonly synthesize agent behaviors by either replaying recorded driving logs or using heuristic-based controllers \cite{sumo18, carla17, svl21}. Such behaviors can often be controlled through a user-friendly high-level programming API \cite{sumo18, carla17} or scenario editor UI \cite{svl21}. However, these methods lack \textit{realism} as they either cannot react to other traffic agents or are not expressive enough to appear human-like.
To address these issues, recent works propose to learn generative models of traffic behavior from large-scale driving datasets \cite{Suo_2021_CVPR, bits2022}. 
While generated behaviors from these models appear reasonable, they merely reflect the distribution of the training data: they lack any mechanism to \emph{control} the generated traffic flow, making these methods less useful in practice.
For example, to test an AV against cut-in from a neighboring lane, a user needs an interface to steer the vehicle to perform the maneuver. Unfortunately, the opaque nature of neural network models makes this difficult for current approaches.

We look to bridge the gap between realism and controllability by studying the problem of \textit{controllable traffic behavior generation}. 
We seek a model that can generate trajectories to meet specific desirable objectives from a user at inference time, which may differ from objectives used in training. 
This is different from traffic models that aim to learn maximum-likelihood behaviors, and therefore are not flexible to new objectives. %
Moreover, controllable generation is an especially challenging for learned approaches, which typically struggle to produce samples outside of the training distribution.

To enable such flexible generation, we leverage recent advances in diffusion modeling, which have achieved state-of-the-art performance in several domains including images\cite{NEURIPS2020_4c5bcfec, dhariwal2021diffusion}, audio\cite{chen2021wavegrad, kong2021diffwave, pmlr-v139-popov21a}, and pedestrian trajectories\cite{Gu_2022_CVPR}. 
Importantly, diffusion models allow a notion of control at generation time through so-called \emph{guidance}, which has benefited several tasks including conditional image generation~\cite{dhariwal2021diffusion}, language generation\cite{li2022nlpdiffusion}, and offline reinforcement learning\cite{janner2022diffuser}.
Inspired by these works, we propose a conditional agent-centric diffusion model (\cref{fig:overview}) that allows flexible traffic behavior generation. 
Unlike prior diffusion models for trajectories\cite{Gu_2022_CVPR,janner2022diffuser}, our model is conditioned on a holistic context surrounding the vehicle including the local roadmap and neighboring agents.
Since traffic has an inherent ground-truth transition function, we also enforce realistic vehicle dynamics into the design of the diffusion model state space to guarantee that generated trajectories are physically feasible. 

To achieve \emph{controllable} generation, Diffuser\cite{janner2022diffuser} \textit{guides} each step of the denoising process in a diffusion model by perturbing network outputs with the gradient of some differentiable objective to encourage desired properties.
For traffic simulation, however, deriving and implementing objectives like collision avoidance, goal reaching, and road rules is complex due to its spatio-temporal and multi-agent nature.
We propose to leverage the established syntax of Signal Temporal Logic (STL) \cite{stl04}. 
As a formal language designed for specifying spatio-temporal constraints, STL allows one to easily and scalably define driving rules; it also incorporates a notion of \textit{robustness} that measures how well rules are satisfied.
Concretely, we use this measure of rule satisfaction as the objective function for guiding the diffusion model (\cref{fig:overview}, right) by leveraging differentiable frameworks\cite{stlrobust21,NEURIPS2019_9015} to make STL compatible with guidance. %
Since our diffusion model generates trajectories independently for each agent in a scene, we further propose a joint guidance procedure for rules involving multi-agent interactions (\textit{e.g.}, no collisions), which simultaneously denoises all agents in the scene to mitigate interaction rule violations.

We evaluate Controllable Traffic Generation (\method), on the  nuScenes \cite{nuscenes} driving dataset, demonstrating the ability to meet user constraints while maintaining realistic trajectory generation. 
In summary, we contribute \textbf{(1)} the problem formulation of controllable traffic generation, \textbf{(2)} a conditional diffusion-based method to generate realistic traffic satisfying physical feasibility and  user-specified STL rules, and \textbf{(3)} extensive evaluation comparing \method to several strong baselines, demonstrating its superiority in terms of the trade-off between controllability and realism.
\textbf{Demos can be found at \url{https://aiasd.github.io/ctg.github.io}}.

\section{Related Work and Background}
\subsection{Traffic Simulation}

Traffic simulation approaches can be categorized into rule-based and learning-based. Rule-based approaches rely on analytical models such as cellular automata and intelligent driver model\cite{microtrafficmodelbenchmark03}. 
These approaches typically have fixed routes for vehicles to follow and separate longitudinal and lateral motions of agents, thus having limited expressiveness for behavior simulation. Learning-based approaches mimic real-world driving behavior based on trajectory datasets~\cite{chai2019multipath,Chen_2022_CVPR,salzmann2020trajectron++}. For example, TrafficSim \cite{Suo_2021_CVPR} uses a trajectory prediction model to perform scene-level traffic simulation. BITS \cite{bits2022} decouples the problem into a high-level latent inference and a low-level driving behavior imitation. However, none of these methods allow a user to specify customized properties of the generated traffic behaviors at inference time. 

Recent work in adversarial or safety-critical scenario generation can be seen as one instance of controllable traffic simulation, namely generating trajectories that cause an AV to misbehave in some way~\cite{wang2021advsim}. STRIVE \cite{rempe2022strive} generates near-collision scenarios by searching in the latent space of a trained trajectory prediction model via a test-time optimization.
Abeysirigoonawardena et al.\cite{Abeysirigoonawardena2019Generating} and Chen et al.\cite{chenbaiming2020} use Bayesian optimization and reinforcement learning, respectively, to generate adversarial trajectories for vehicles in a specific scenario like intersection crossing or lane changing.
While these works focus specifically on adversarial objectives, our approach is general and can generate trajectories to meet several different objectives.

\subsection{Diffusion Modeling}
Controllable diffusion models have been explored with classifier~\cite{dhariwal2021diffusion}, classifier-free~\cite{ho2022classifier}, and reconstruction~\cite{ho2022video} guidance for image and video generation. 
Li et al.\cite{li2022nlpdiffusion} use a diffusion model with different pre-trained classifiers to guide language generation on different natural language tasks. 
Diffuser \cite{janner2022diffuser} uses a diffusion model to plan robot behavior (state-action trajectories). Gu et al.\cite{Gu_2022_CVPR} model pedestrian trajectories for forecasting.
Different from these works that use no or limited context, we adapt conditional diffusion to condition on decision-relevant information such as map and state of nearby agents.
Additionally, we leverage known vehicle dynamics models to ensure physical feasibility of the generated trajectories. Diffuser\cite{janner2022diffuser} also introduces test-time guidance to generate trajectories that optimize a given reward function. We build on this formulation to generate controllable traffic trajectories, but rather than learning reward functions we use analytical loss functions based on STL rules that are easy and scalable for driving applications.

\subsection{Signal Temporal Logic (STL)}
\label{sec:stl}
STL \cite{stl04} formulas are interpreted over signals, $\btau{} = s_t,...,s_{t+T}$, an ordered finite sequence of states $s_i\in\mathbb{R}^n$. 
A signal represents a sequence of real-valued, discrete-time outputs from a system. 
STL formulas are recursively defined based on the following context-free grammar:
\begin{align}
\phi::= &~~\top ~|~ \mu_c ~|~ \neg\phi ~|~ \phi \wedge \psi ~|~  \phi\,\mathcal{U}_{[a,b]}\,\psi.
\label{eq:STL grammar}
\end{align}
\noindent The grammar defines a list of expressions, each separated by the pipe ( $|$ ), that can be used to construct an STL formula. In particular, a formula $\phi$ is generated by selecting expressions recursively. We briefly summarize \cref{eq:STL grammar} here, but refer readers to Leung et al. \cite{stlrobust21} (Section 2.2) for a pedagogical introduction to STL.
The core of an STL formula are predicates $\mu_c$ of the form $\mu(z) > c$, where $c\in\mathbb{R}$ and $\mu: \mathbb{R}^n \rightarrow \mathbb{R}$ is a differentiable function. These define the desired properties/constraints of a signal. 
Additionally, $\top$ means true, $\phi$ and $\psi$ are STL formulas, and $[a,b]\subseteq \mathbb{R}_{\geq 0}$ is a time interval (assumed to be $[0, \infty )$ if omitted).
The symbol $\neg$ is negation, while other symbols describe how to combine multiple formulas: $\wedge$ (conjunction/and) and $\mathcal{U}$ (until), which is a temporal operator.
Other common logical connectives ($\vee$ disjunction/or, $\Rightarrow$ implies), and temporal operators ($\lozenge$ eventually, $\square$ always) can be derived from \cref{eq:STL grammar}. 

Importantly, STL includes a formal notion of robustness, which measures how much a signal satisfies or violates a formula~\cite{stl04}. We use \textit{robustness formulas} as guidance functions for the proposed conditional diffusion model.
In contrast to Leung and Pavone \cite{leung2022} that %
focus on a single-controller setting and rules fixed at training time, we consider a more flexible inference-time guidance with multiple agents.

\section{Controllable Traffic Generation}
\label{sec:method}
Next, we detail our approach \method. \cref{sec:method-formulation} formulates the problem of controllable traffic generation. We then describe the two stages of \method.
The offline stage trains a dynamics-enforced conditional diffusion model to capture diverse behaviors from real-world driving data (\cref{sec:method-diffuser}).
Then during online inference, \method generates rule-compliant behaviors by sampling the model using a novel iterative joint STL guidance process (\cref{sec:method-guidance}).
Taken together, the conditional diffusion model and STL-based guidance enable realistic and controllable generation of traffic trajectories.

\subsection{Problem Formulation}
\label{sec:method-formulation}
For some target vehicle (tgt) that we would like to simulate, let the state at a timestep $t$ be $s_{t}^{\text{tgt}}=(x_{t}^{\text{tgt}},y_{t}^{\text{tgt}}, v_{t}^{\text{tgt}}, \theta_{t}^{\text{tgt}})$, including 2D location, speed, and yaw. Similarly, let the action (\textit{i.e.}, control) be $a_{t}^{\text{tgt}}=(\dot{v}_{t}^{\text{tgt}},\dot{\theta}_{t}^{\text{tgt}})$ with acceleration and yaw rate. We denote $\bc=(I,S)$ to be decision-relevant context for the target agent. This consists of a local (agent-centric) semantic map $I$ and the $H$ previous states of both the target agent and its $M$ neighbors $S_{t-H:t}=\{s_{t-H:t}^{\text{tgt}},s_{t-H:t}^1, \dots, s_{t-H:t}^M\}$.
To obtain state $s_{t+1}^m$ for vehicle $m$ at time $t+1$, we assume a transition function $f$ that computes $s_{t+1}^m=f(s_{t}^m, a_{t}^m)$ given the previous state $s_{t}^m$ and control $a_{t}^m$. We use a unicycle dynamics model for $f$. 

Our goal is to generate realistic and rule-satisfying traffic behavior for the target agent given (1) the decision context $\bc$ and (2) a function  $r : \mathbb{R}^{4T} \times \mathbb{R}^{2T} \rightarrow \mathbb{R}$ to measure rule satisfaction of a state and action trajectory.
A model should generate a future trajectory for the target agent $s_{t:t+T}^{\text{tgt}}$ over the next $T$ time steps.
Ideally, this trajectory maximizes satisfaction $r(s_{t:t+T}^{\text{tgt}}, a_{t:t+T}^{\text{tgt}})$ to avoid violating the given rule.
However, in many cases there is an inherent tradeoff between rule satisfaction and trajectory realism: \textit{e.g.}, if a user seeks a simulated vehicle with a speed much slower/faster than the speed limit, this is naturally ``unrealistic" in the context of city streets. 
Therefore we must strike a balance between meeting user-specified constraints while still maintaining realistic behavior.
As described next, our method does this by first training a rule-agnostic traffic generation model on real-world data to capture \textit{realism}, which is then guided for rule-specific compliance only during inference. 

\subsection{Conditional Diffusion for Traffic Modeling}
\label{sec:method-diffuser}
\begin{figure}[t]
\centering
\includegraphics[width=0.425\textwidth]{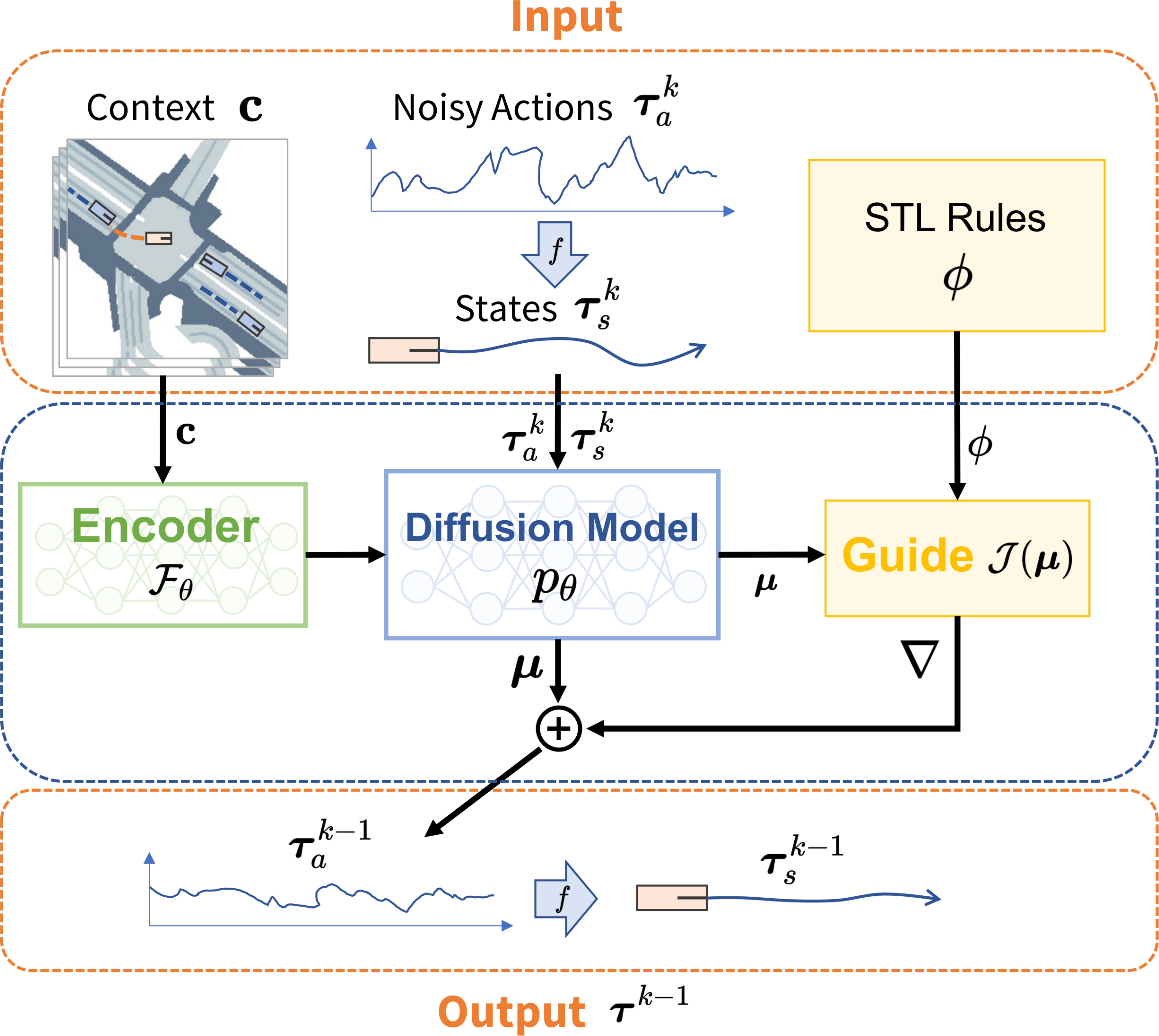}
\caption{\textbf{\small{Single test-time denoising step.}} \small{Given current noisy actions and states along with encoded context, the diffusion model predicts the mean of the next-step distribution which is then perturbed according to the desired guidance function.}}
\label{fig:state_action}
\vspace{-5mm}
\end{figure}

Diffusion models \cite{pmlr-v37-sohl-dickstein15,NEURIPS2020_4c5bcfec} pose data generation as an iterative denoising process by learning to reverse a forward diffusion process.
As shown in \cref{fig:state_action}, our diffusion model operates primarily on a (future) trajectory of states and actions~\cite{janner2022diffuser}, but is conditional as it receives the context as input at each step of denoising. 
Starting from Gaussian noise, the diffusion model is applied iteratively to predict a clean, denoised trajectory of states and actions.

\noindent\textbf{Trajectory representation.} 
In this section, we denote the (future) trajectory that the model operates on as:
\begin{equation*}
    \btau{} := \begin{bmatrix}
            \btau{}_a  \\
            \btau{}_s  \\
        \end{bmatrix}, \quad 
        \btau{}_a := [a_0~...~a_{T-1}], \quad 
        \btau{}_s := [s_1~...~s_{T}].
\end{equation*}
Unlike \cite{janner2022diffuser} which directly predicts states and actions jointly, our model only predicts actions $\btau{}_a$ and we leverage the known dynamics $f$ to infer states $\btau{}_s$ via rollout starting at the initial state $s_0$ (included in the past context). In other words, in the following formulation, $\btau{}_s$ always refers to a state trajectory resulting from actions, or, more formally: $\btau{}_s = f(s_0, \btau{}_a)$. This ensures physical feasibility of the state trajectory throughout the denoising process.

\noindent\textbf{Formulation.}  
Let $\btau{k}_a$ be the action trajectory at the $k$th diffusion step where $k=0$ is at the original clean trajectory.
The forward diffusion process acting on $\btau{0}_a$ is defined as: %
\begin{equation}
\setlength{\abovedisplayskip}{1pt}
\setlength{\belowdisplayskip}{1pt}
\begin{aligned}
\label{eq: diffusion_q} q(\btau{1:K}_a|\btau{0}_a) &:=  \prod_{k=1}^{K} q(\btau{k}_a|\btau{k-1}_a) \\
q(\btau{k}_a|\btau{k-1}_a) &:= \mathcal{N}(\btau{k}_a; \sqrt{1-\beta_{k}}\btau{k-1}_a,\beta_{k}\mathbf{I}),
\end{aligned}
\end{equation}
where $\beta_{1}, \beta_{2}, \cdots \beta_{K}$ are a fixed variance schedule that controls the scale of the injected noise at each diffusion step. 
As noise is gradually added, the signal is corrupted into an isotropic Gaussian distribution. 

For trajectory generation, we seek to reverse this diffusion process using a learned conditional denoising model (\cref{fig:state_action}) that is iteratively applied starting from sampled noise.
Given the context information $\bc$, the reverse diffusion process is:
\begin{equation}
\setlength{\abovedisplayskip}{1pt}
\setlength{\belowdisplayskip}{1pt}
\begin{aligned}
\label{eq: denoise_p} p_{\theta}(\btau{0:K}_a | \bc) &:=  p(\btau{K}_a)\prod_{k=1}^{K} p_{\theta}(\btau{k-1}_a|\btau{k}, \bc) \\
 p_{\theta}(\btau{k-1}_a|\btau{k}, \bc) &:= \mathcal{N}(\btau{k-1}_a; \boldsymbol{\mu}_{\theta}(\btau{k},k,\bc),\boldsymbol{\Sigma}_{\theta}(\btau{k},k,\bc)), \\
\end{aligned}
\end{equation}
where $p(\btau{K}_a) = \mathcal{N}(\mathbf{0}, \mathbf{I})$ and $\theta$ denotes the parameters of the diffusion model. 
Note that at each step, the model receives \textit{both} actions $\btau{k}_a$ and the resulting states $\btau{k}_s = f(s_0, \btau{k}_a)$ as input. 
Following ~\cite{NEURIPS2020_4c5bcfec, Gu_2022_CVPR}, the variance term of the Gaussian transition is fixed as $\boldsymbol{\Sigma}^k = \boldsymbol{\Sigma}_{\theta}(\btau{k},k,\bc) = \sigma_{k}^2\mathbf{I} = \beta_{k}\mathbf{I}$.

\noindent\textbf{Training.}
At each training iteration, the context $\mathbf{c}$ and ground truth clean trajectory $\btau{0}$ are sampled from a real-world driving dataset and the denoising step $k$ is uniformly sampled from $\{1, \dots, K\}$.
We compute the noisy input $\btau{k}$ from $\btau{0}$ by first corrupting the action trajectory $\btau{k}_a = \sqrt{\bar{\alpha}_{k}} \btau{0}_a + \sqrt{1-\bar{\alpha}_{k}} \epsilon $, $\epsilon \sim \mathcal{N}(0,\mathbf{I})$ with $\bar{\alpha}_{k} = \prod_{l=0}^k 1 - \beta_l$, and then computing the corresponding state $\btau{k}_s = f(s_0, \btau{k}_a)$.
The diffusion model indirectly parameterizes $\boldsymbol{\mu}_\theta$ in \cref{eq: denoise_p} by instead predicting the uncorrupted trajectory $\hat{\boldsymbol{\tau}}^0 = [ \hat{\boldsymbol{\tau}}^0_a; \; f(s_0, \hat{\boldsymbol{\tau}}^0_a)]$ where $\hat{\boldsymbol{\tau}}^0_a = \hat{\boldsymbol{\tau}}^0_a(\btau{k},k,\bc{})$ is the direct network output (see \cite{li2022nlpdiffusion,janner2022diffuser,nichol2021improved}). Finally, we use a simplified loss function to train the model:
\begin{equation}
\begin{aligned}
\label{eq: loss_simple} 
L(\theta) = \mathbb{E}_{\epsilon,k, \btau{0}, \bc} \left[ || \btau{0} - \hat{\boldsymbol{\tau}}^0 ||^2 \right].
\end{aligned}
\end{equation}
Note that both the action and state trajectories are supervised by this loss since we found that the additional state trajectory information improves generation quality.

\noindent\textbf{Implementation details.} The input context $\mathbf{c}$ containing agent-centric map information and past trajectories is represented in a rasterized format. This context is processed by a ResNet~\cite{he2016deep} encoder $\mathcal{F_{\theta}}$ before being passed to the diffusion model. 
Similar to Diffuser \cite{janner2022diffuser}, we use a diffusion model architecture like U-Net containing several blocks of temporal 1D convolutions over the input trajectory. 
We incorporate conditioning information by first concatenating with the diffusion step input $k$ and then adding this conditioning feature to the convolutional features at each block of the U-Net.
The diffusion process uses a cosine variance schedule~\cite{janner2022diffuser,nichol2021improved} and $K=100$ diffusion steps for all experiments.

\subsection{Guided Generation with Signal Temporal Logic (STL)}
\label{sec:method-guidance}
To enforce desired rules on realistic samples from the trained diffusion model, we introduce an iterative guidance algorithm with rules specified as STL formulas.

\noindent\textbf{Conditional guidance formulation.}
Diffuser~\cite{janner2022diffuser} introduces the notion of \textit{guidance} to sample trajectories from an \textit{unconditional} diffusion model to meet some pre-defined objective. We can do the same for our conditional diffusion model by defining a binary random variable $\mathcal{O}$ that indicates if a trajectory is optimal, with ${p(\mathcal{O}=1) = \exp(r(\btau{}_s, \btau{}_a))}$ based on the rule satisfaction ``reward" $r$. 
To approximately sample from the distribution of optimal trajectories, the steps of the denoising process can be modified to~\cite{pmlr-v37-sohl-dickstein15,janner2022diffuser}:
\begin{equation}
\label{eq:guided}
p_\theta(\btau{k-1}_a \mid \btau{k}, \bc, \opt) \approx \mathcal{N}(\btau{k-1}_a; \boldsymbol{\mu} + \boldsymbol{\Sigma} g, \boldsymbol{\Sigma}),
\end{equation}
where $\boldsymbol{\mu} = \bmu_\theta, \; \boldsymbol{\Sigma} = \bSigma_\theta$ as in \cref{eq: denoise_p}, and the added gradient is computed from a \textit{guide} $\mathcal{J}$ based on satisfaction: 
\begin{align}
g &= \nabla \mathcal{J}(\boldsymbol{\mu}) = \nabla_{\boldsymbol{\mu}}\; r(f(s_0, \boldsymbol{\mu}),\boldsymbol{\mu}).
\end{align}

The process of perturbing the predicted means from the diffusion model using gradients of a specified objective is summarized in \cref{alg:sampling}. 
Different from Diffuser \cite{janner2022diffuser}, an iterative inner gradient descent with clipping using the Adam optimizer is incorporated rather than using a single-step gradient update.
This gives flexibility to trade off rule compliance and realism by adjusting learning rate and the number of optimization steps.
When generating a future trajectory in practice, we guide several samples from the diffusion model and choose the one with the best rule satisfaction according to $\mathcal{J}$ at the end of denoising. We refer to this as \textit{filtration}.

{\centering
\begin{figure}[t]
\vspace{-2mm}
\begin{minipage}{\linewidth}
  \begin{algorithm}[H]
  \scriptsize
    \caption{Guided Sampling}
    \label{alg:sampling}
    \begin{algorithmic}[1]
    \STATE \textbf{Require} encoder $\mathcal{F_{\theta}}$, conditional diffusion model $\boldsymbol{\mu}_\theta$, transition function $f$, guide $\mathcal{J}$, scale $\alpha$, covariances $\boldsymbol{\Sigma}^k$, diffusion steps $K$, inner gradient descent steps $M$, number of actions to take before re-planning $l$. \\
    \WHILE{not done}
        \STATE Observe state $s_0$ and context $\bc$
        \STATE Initialize trajectory $\btau{K}_a \sim \mathcal{N}(\bm{0}, \bm{I})$; $\btau{K}_s = f(s_0, \btau{K}_a)$; $\btau{K}=[\btau{K}_a; \btau{K}_s]$
        \FOR{$k = K, \ldots, 1$}
            \STATE $\bmu = \bmu_\theta(\btau{k}, k, \mathcal{F_{\theta}}(\bc))$ \\
            \STATE $\bmu^{(0)} = \bmu$\\
            \FOR{$j = 1, \ldots, M$}
                \STATE $\bmu^{(j)} = \bmu^{(j-1)} + \alpha \nabla \mathcal{J}(\bmu^{(j-1)})$\\
                \STATE $\Delta \bmu = |\bmu^{(j)}-\bmu^{(0)}|$; $\Delta \bmu \gets \textrm{clip}(\Delta \bmu, -\beta_k, \beta_k)$\\
                \STATE $\bmu^{(j)} \gets \bmu^{(0)} + \Delta \bmu$ \\
            \ENDFOR
            \STATE $\btau{k-1}_a \sim \mathcal{N}(\bmu^{(M)}, \bSigma^k)$; $\btau{k-1}_s = f(s_0, \btau{k-1}_a)$;\\ $\btau{k-1}=[\btau{k-1}_a; \btau{k-1}_s]$\\
        \ENDFOR
        \STATE Execute first $l$ actions of trajectory $\btau{0}_{a}$
    \ENDWHILE
    \end{algorithmic}
  \end{algorithm}
\end{minipage}
\vspace{-6mm}
\end{figure}
}

\noindent\textbf{STL as guidance.}
Instead of training a classifier or reward function for $\mathcal{J}$ as in prior works \cite{janner2022diffuser, li2022nlpdiffusion}, our guidance functions are implemented analytically based on STL.
For each rule we wish to apply through guidance, the grammar in \cref{eq:STL grammar} is used to generate a corresponding STL formula describing how the trajectory should be constrained.
Examples of various STL rules used are shown in \cref{table:stl}; note the formulas are relatively simple despite the complex behavior they describe.
By construction, every STL formula admits a \textit{robustness formula} measuring the degree of rule satisfaction, which is used as the guide  $\mathcal{J}$.
Since it is necessary to compute a gradient through this function, STL formulas are implemented using differentiable frameworks\cite{stlrobust21,NEURIPS2019_9015}.

\noindent\textbf{Multi-agent guidance.}
A particular challenge is applying \textit{scene-level} rules that involve multiple agents (\textit{e.g.}, no collisions). For this purpose, guided sampling is performed in a batched fashion over all agents in the same scene simultaneously. This way guidance can be computed across all trajectories, and the corresponding gradients are collected and propagated back to agents as needed.

\noindent\textbf{Simulating traffic.}
To perform closed-loop traffic simulation of a scene with many agents, the same model is used for each agent in a standard control loop: for each agent at each step of simulation, a guided sample is generated from the model and the first few actions are taken before re-planning at a prescribed frequency. In all experiments (\cref{sec:experiments}), each scene is rolled out for 20 seconds starting from a ground truth driving log, and the re-plan rate is 2 Hz as in \cite{bits2022}.

\section{Experiments}
\label{sec:experiments}

\begin{table*}[ht]
\centering
\scriptsize
\caption{
    \small{Definition of rules. For each rule, the STL formula and corresponding metric used for evaluation are shown.} \label{table:stl}
}
\resizebox{.96\textwidth}{!}{
\begin{tabularx}{1.05\textwidth}{l|l|l}
\toprule
Rule & STL Formula & Evaluation Metric  \\ 
\midrule
 speed limit & $\phi_{\textrm{speedlimit}}:=\bigwedge_{i} \square (v_{i,t}-v_{\textrm{limit}_i}) < \epsilon$ & $h_{\textrm{speedlimit}}:=\Sigma_i \Sigma_t \max (0, v_{i,t}-v_{\textrm{limit}_i} ) $ \\ 
target speed & $\phi_{\textrm{targetspeed}}:=\bigwedge_{i,t} \vert v_{i,t}-v_{\textrm{target}_{i,t}} \vert < \epsilon$ & $h_{\textrm{targetspeed}}:=\Sigma_i \Sigma_t  \vert v_{i,t}-v_{\textrm{target}_{i,t}} \vert $ \\
 no collision  & $\phi_{\textrm{collision}}:=\square \bigwedge_{i\neq j} \vert\vert (x_{i,t}, y_{i,t}) - (x_{j,t}, y_{j,t}) \vert\vert > \epsilon$ & $h_{\textrm{collision}}:=\Sigma_{i\neq j} \mathds{1}[\vert\vert (x_{i,t}, y_{i,t}) - (x_{j,t}, y_{j,t}) \vert\vert \leq \epsilon]$ \\
 no off-road  & $\phi_{\textrm{offroad}}:=\bigwedge_{i,w} \square  \vert\vert (x_{i,t}, y_{i,t}) - (x_{\textrm{off-road}_w}, y_{\textrm{off-road}_w}) \vert\vert > \epsilon$ & $h_{\textrm{offroad}}:=\Sigma_i\mathds{1}[\min_{w} \vert\vert (x_{i,t}, y_{i,t}) - (x_{\textrm{off-road}_w}, y_{\textrm{off-road}_w}) \vert\vert \leq \epsilon]$ \\
 goal waypoint & $\phi_{\textrm{waypoint}}:=\bigwedge_{i} \lozenge \vert\vert (x_{i,t}, y_{i,t}) - (x_{\textrm{goal}_i}, y_{\textrm{goal}_i}) \vert\vert < \epsilon$ & $h_{\textrm{waypoint}}:=\Sigma_i \min_{t} \vert\vert (x_{i,t}, y_{i,t}) - (x_{\textrm{goal}_i}, y_{\textrm{goal}_i}) \vert\vert $ \\
 stop sign  & $\phi_{\textrm{stopsign}}:=\bigwedge_{i} \square (\phi_{i~\textrm{in stop region}} \Rightarrow \lozenge \square_{[0,m]} (\phi_{i~\textrm{in stop region}} \wedge \phi_{\textrm{stop}}) )$ & $h_{\textrm{stopsign}}:=\Sigma_i \min_{i~\textrm{in stop region}} |v_{i,t}|$ \\ 
stop sign + no off-road  & $\phi_{\textrm{stopsign}} \wedge \phi_{\textrm{offroad}}$ & $h_{\textrm{stopsign}},h_{\textrm{offroad}}$ \\
 goal waypoint + target speed & $\phi_{\textrm{waypoint}} \wedge \phi_{\textrm{targetspeed}}$ & $h_{\textrm{waypoint}},h_{\textrm{targetspeed}}$ \\
\bottomrule
\end{tabularx}
}
\end{table*}

\begin{table*}[ht]
\centering
\scriptsize
\caption{
    \small{Quantitative results (Single Rule). The top two methods for each metric are highlighted.} \label{table:main_result_nuscenes_single_rule}
}
\begin{tabularx}{0.94\textwidth}{l|ccc|ccc|ccc|ccc|ccc}
\toprule
                 & \multicolumn{3}{c|}{\textbf{speed limit}} & \multicolumn{3}{c|}{\textbf{target speed}} &  \multicolumn{3}{c|}{\textbf{no collision}} & \multicolumn{3}{c|}{\textbf{no off-road}} & \multicolumn{3}{c}{\textbf{goal waypoint}} \\
& rule             & real  & fail  & rule  & real  & fail  & rule  & real  & fail  &
rule  & real  & fail  & rule  & real  & fail     \\
\midrule
SimNet                          & 0.739          & 0.898          & 0.353          & 1.989          & 0.898          & 0.353          & 0.137          & 0.898          & 0.353          & 0.453          & 0.900          & 0.353          & 7.543          & 0.900          & 0.353          \\
SimNet+opt                      & 0.038          & 0.770          & 0.470          & 0.630          & 1.234          & 0.593          & 0.045          & 1.149          & 0.398          & 0.427          & 1.242          & 0.416          & 1.947          & 1.162          & 0.467          \\
TrafficSim                      & 0.737          & 1.362          & 0.443          & 1.922          & 1.346          & 0.440          & 0.140          & 1.542          & 0.416          & 0.485          & 1.564          & 0.401          & 7.733          & 1.574          & 0.458          \\
TrafficSim+opt                  & 0.042          & 1.444          & 0.325          & \textbf{0.610} & 1.634          & 0.404          & 0.075          & 1.063          & \textbf{0.265} & 0.423          & 1.836          & 0.413          & 2.414          & 1.766          & 0.532          \\
BITS                            & 0.188          & 1.068          & \textbf{0.256} & 1.054          & 0.968          & \textbf{0.246} & \textbf{0.038} & 1.220          & 0.314          & 0.432          & 1.131          & \textbf{0.296} & 4.493          & 1.152          & \textbf{0.332} \\
BITS+opt                        & \textbf{0.033} & 1.190          & 0.487          & 0.681          & 1.434          & 0.542          & \textbf{0.028} & 1.617          & 0.354          & 0.435          & 1.261          & 0.343          & \textbf{1.533} & 1.280          & 0.442          \\
\method(w/o f+g) & 1.380          & \textbf{0.396} & 0.301          & 2.662          & \textbf{0.396} & 0.301          & 0.172          & \textbf{0.396} & 0.301          & \textbf{0.369} & \textbf{0.396} & \textbf{0.301} & 7.052          & \textbf{0.396} & \textbf{0.301} \\
\method          & \textbf{0.019} & \textbf{0.359} & \textbf{0.165} & \textbf{0.150} & \textbf{0.855} & \textbf{0.179} & 0.040          & \textbf{0.569} & \textbf{0.271} & \textbf{0.341} & \textbf{0.501} & 0.455          & \textbf{1.943} & \textbf{0.564} & 0.387     \\
\bottomrule
\end{tabularx}
\vspace{-4mm} 
\end{table*}

\begin{table}[ht]
\centering
\scriptsize
\caption{
    \small{Quantitative results (Multiple Rules).} \label{table:main_result_nuscenes_multi_rule}
}
\resizebox{.48\textwidth}{!}{
\begin{tabularx}{0.55\textwidth}{l|cccc|cccc}
\toprule
                 & \multicolumn{4}{c|}{\textbf{stop sign + no off-road}} & \multicolumn{4}{c}{\textbf{goal waypoint + target speed}}  \\
                 & rule1 & rule2    & real    & fail    & rule1 & rule2    & real    & fail           \\
\midrule 
SimNet                          & 2.282          & 0.454          & 0.898          & 0.353          & 3.803          & 1.610          & 0.898          & 0.353          \\
SimNet+opt                      & 1.527          & 0.480          & 0.796          & 0.443          & 3.238          & 0.980          & 1.189          & 0.659          \\
TrafficSim                      & 2.670 & 0.484          & 1.605          & 0.409          & \textbf{1.583} & 4.544          & 1.369          & 0.450          \\
TrafficSim+opt                  & \textbf{0.849}          & 0.405          & 1.577          & \textbf{0.281} & 2.817          & \textbf{0.995} & 1.398          & 0.529          \\
BITS                            & 2.023          & 0.434          & 1.032          & \textbf{0.254} & 2.677          & 1.077          & 0.919          & \textbf{0.286} \\
BITS+opt                        & 1.299          & 0.471          & 1.083          & 0.399          & 3.171          & 1.019          & 1.455          & 0.601          \\
\method(w/o f+g) & 2.573          & \textbf{0.369} & \textbf{0.396} & 0.301          & 3.868          & 2.202          & \textbf{0.396} & \textbf{0.301} \\
\method          & \textbf{0.528} & \textbf{0.338} & \textbf{0.288} & 0.326          & \textbf{1.205} & \textbf{0.231} & \textbf{0.738} & 0.329          \\
\bottomrule
\end{tabularx}
}
\vspace{-3mm}
\end{table}

We conduct experiments to validate that: (1) \method can generate controllable traffic behaviors that satisfy user-specified rules, and (2) compared to strong baselines, \method achieves better rule satisfaction while maintaining realism.
As discussed in \cref{sec:method-formulation}, there is often a trade-off between realism and rule compliance; ideally a method will strike a reasonable balance and achieve good performance for both.
After describing the experimental design (\cref{sec:setup}), we compare to baselines in single-rule (\cref{sec:quantitative_single}) and multi-rule (\cref{sec:quantitative_multi}) settings both quantitatively and qualitatively, and finally conduct an ablation study (\cref{sec:ablation}).

\subsection{Experimental Setup}
\label{sec:setup}
\noindent\textbf{Datasets.} 
nuScenes \cite{nuscenes} is a large-scale real-world driving dataset, which consists of 5.5 hours of accurate trajectories across two cities with diverse scenarios and dense traffic.
We train all models on scenes from the train split and evaluate on 100 scenes randomly sampled from the validation split. In the current work, we focus only on vehicle simulation and defer other types (\textit{e.g.}, pedestrians, cyclists) to future works.

\noindent\textbf{Metrics.}
Our evaluation focuses on controllability, realism, and stability (\textit{i.e.}, avoiding collisions and off-road driving). 
We consider rule-specific violation metrics (\textbf{rule}), detailed in \cref{table:stl}, to evaluate \textit{controllability}. Metrics are computed for each scene ($i$ denotes each vehicle in the scene), then averaged across all testing scenes. 
To evaluate \textit{realism}, we follow \cite{bits2022} and compare data statistics between generated traffic simulations and ground truth trajectories in the dataset. 
This comparison is computed via the Wasserstein distance between the normalized histograms of the driving profiles for the simulated and recorded trajectories. 
We define an aggregated metric -- realism deviation (\textbf{real}) -- as the mean of the realism for three properties from \cite{bits2022}: longitudinal acceleration magnitude, latitudinal acceleration magnitude, and jerk. 
We further provide failure rate (\textbf{fail}) to evaluate the \textit{stability} of generated trajectories. This is measured as the average fraction of agents experiencing a critical failure, \textit{i.e.} collision or road departure, in a scene.

\noindent\textbf{Baselines.} 
Since there are no comparable works on rule-compliant traffic generation, we augment state-of-the-art traffic simulation models by adding a test-time optimization to meet specified rules. For a fair comparison, this optimization uses the same loss function as used for guidance in \method. \textbf{SimNet} \cite{bergamini2021simnet} is a deterministic behavior-cloning model. We apply an optimization on its output action trajectory (\textbf{SimNet+opt}). 
TrafficSim \cite{Suo_2021_CVPR} is a CVAE-based trajectory generation method. We consider a variant with filtration (\textbf{TrafficSim}) using our loss function, and another with both filtration and latent space optimization (\textbf{TrafficSim+opt}). BITS \cite{bits2022} is a bi-level imitation learning model and we adapt its sampling ranking function to use our loss function (\textbf{BITS}). We also use a variant that employs optimization on the output action trajectory (\textbf{BITS+opt}). Finally, we compare to \method \textit{without} filtration and guidance (\textbf{\method w/o f+g}), \textit{i.e.} random samples from the diffusion model.

\subsection{Single Rule Evaluation}
\label{sec:quantitative_single}
We first evaluate how well methods satisfy a single specified rule, which is crucial for applications such as traffic scene editing.
We apply five STL rules formulated in \cref{table:stl}: speed limit, target speed, goal waypoint, no collision, and no off-road. 
For rules that require specific parameters to be set (\textit{e.g.}, the goal waypoint location), we select reasonable values based on the ground truth log in the dataset to avoid setting out-of-distribution values (\textit{e.g.} off-road waypoints).

\noindent\textbf{Speed Limit.} Vehicles should not exceed a speed limit threshold. Since the speed limit of each road is not available in the dataset, we set the limit per scene to be the speed at the $75\%$ quantile of all moving vehicles in that scene.

\noindent\textbf{Target Speed.} Vehicles should follow a specified speed at each time step. For each vehicle, the speed is set to $50\%$ of its speed in the ground truth scene, similar to a traffic jam.

\noindent\textbf{Goal Waypoint.} Vehicles should reach a specified waypoint at any time in the future. We set waypoints to be the position at $15$s along the ground truth data trajectory for each vehicle. Hitting a waypoint from ground truth data is not trivial since sampled trajectories often greatly deviate from the dataset. 

\noindent\textbf{No Collision.} Vehicles should not collide with each other.

\noindent\textbf{No Off-road.} Vehicles should not leave the drivable area.

\begin{figure}[t]
\scriptsize
\centering
\subfloat[][BC+opt(1.3)]{\includegraphics[width=0.15\textwidth]{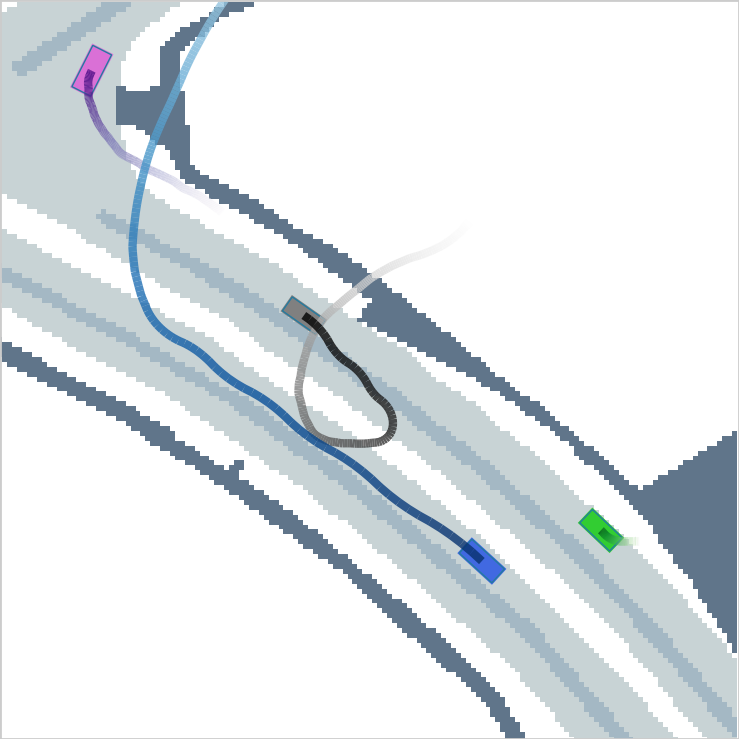}}
\subfloat[][TrafficSim+opt(1.1)]{\includegraphics[width=0.15\textwidth]{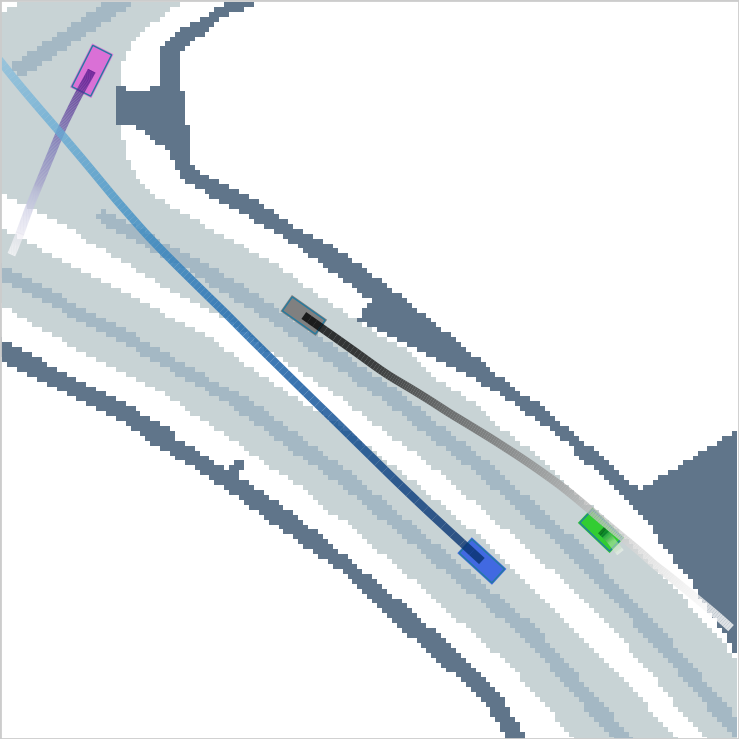}}
\subfloat[][\method{}(0.3)]{\includegraphics[width=0.15\textwidth]{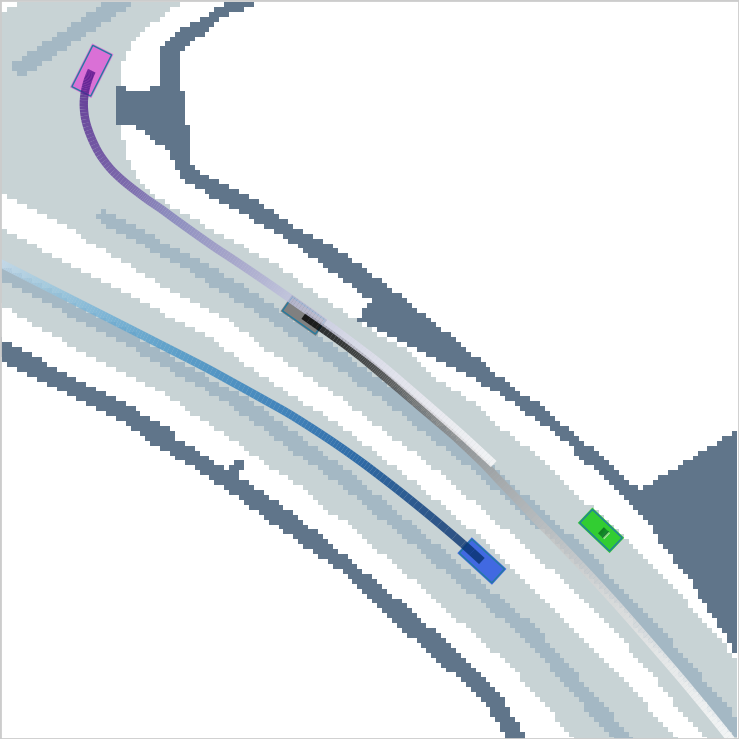}}

\subfloat[][BC+opt(2.15)]{\includegraphics[width=0.15\textwidth]{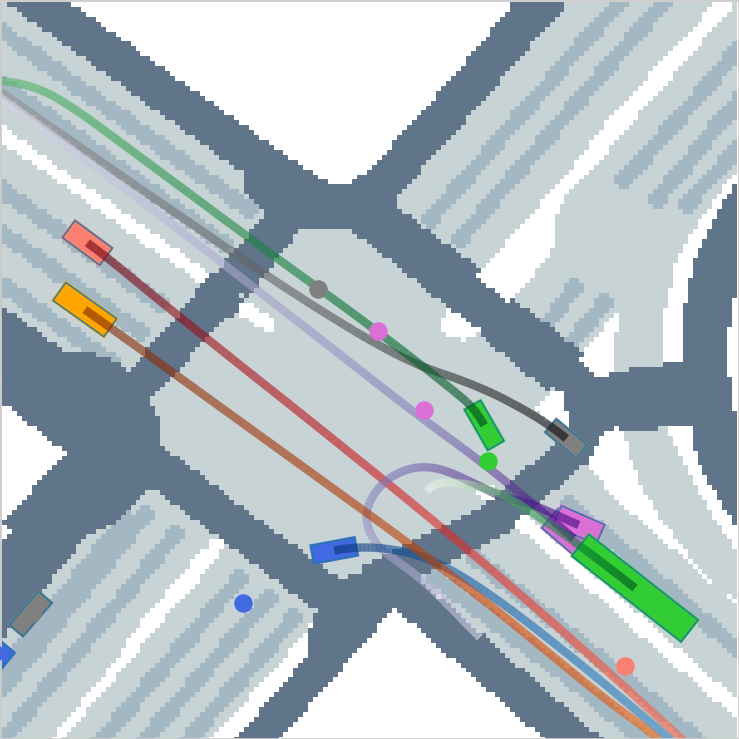}}
\subfloat[][BITS+opt(6.28)]{\includegraphics[width=0.15\textwidth]{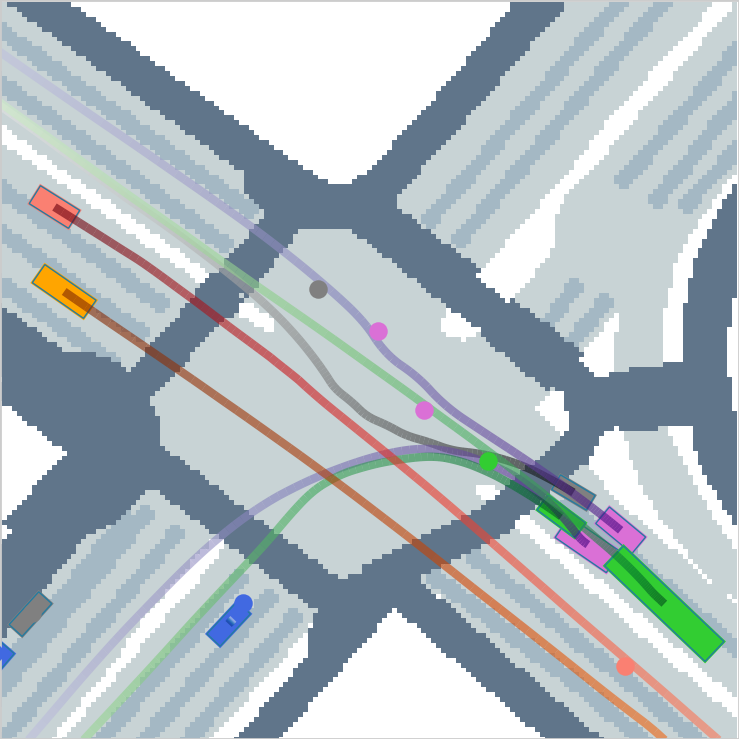}}
\subfloat[][\method{}(3.17)]{\includegraphics[width=0.15\textwidth]{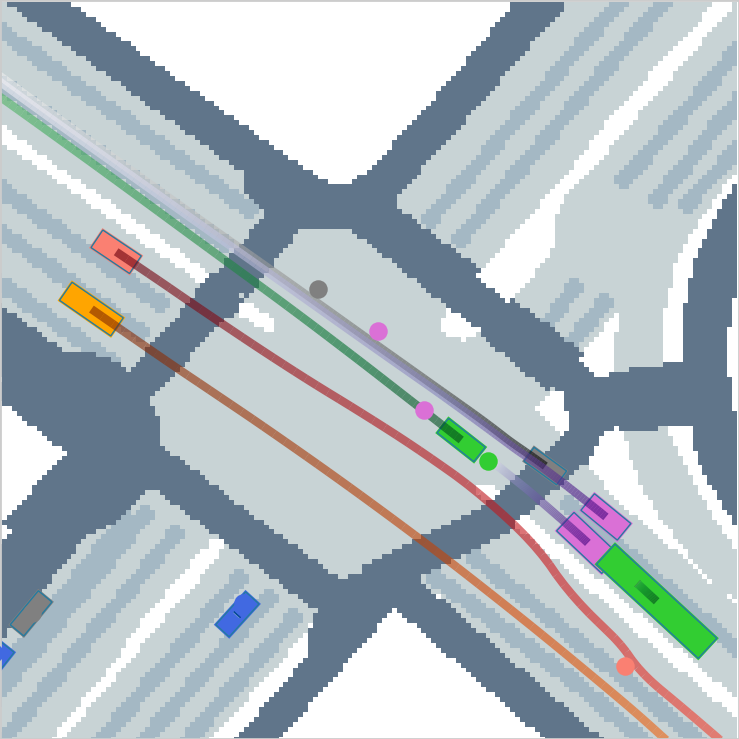}}

\caption{\small{Target speed (a)-(c) and goal waypoint (d)-(f) rule results. Rule violations are in parentheses ($m/s$ and $m$, respectively).}}
\label{fig:example}
\vspace{-4mm}
\end{figure}

Quanitative results are shown in \Cref{table:main_result_nuscenes_single_rule}.
In general, \method achieves lower values for rule violation, realism deviation, and failure rate than the baselines.
Among all five settings, \method has the lowest rule violation in three and is competitive in the others. %
For realism deviation and failure rate, \method is usually top two.
\cref{fig:example} shows qualitative results for target speed and waypoint rules.
Compared to TrafficSim+opt and BC+opt, the strongest baselines for target speed, \method has the lowest rule violation using more realistic trajectories.
For the waypoint example, although BC+opt provides better rule satisfaction than \method, both BC+opt and BITS+opt predict curvy, unrealistic trajectories resulting in multiple collisions. 

\subsection{Multiple Rules Evaluation}
\label{sec:quantitative_multi}
\noindent\textbf{Stop Sign and No Off-road.} Vehicles should stop if they enter a stop sign region and not go off-road.
The expression for this rule (\cref{table:stl}) is relatively involved using ``Implies" and ``Eventually" operators, making it a good test for our STL-based approach.
Stop regions are $20 \times 20$ m boxes with centers set to be $5$s along the ground truth data trajectories.

\noindent\textbf{Goal Waypoint and Target Speed.} Vehicles should reach their goal following the specified target speeds. Waypoints are set $10$s along the ground truth trajectories, and the target speeds at each time step are the same as the ground truth scene.
This setting is similar to a ``reactive replay" use case in AV testing, where a user reconstructs a driving log using the traffic model, which allows agents to realistically react to any subsequent changes to AV behavior or environment.

Results are shown in \Cref{table:main_result_nuscenes_multi_rule}. For both settings, \method variants achieve the top two lowest rule violation and realism deviation, with only slightly higher failure rates.

\subsection{Ablation Study.}
\label{sec:ablation}

\begin{table}[t]
\centering
\scriptsize
\caption{
    \small{Ablation study of CTG features: dyn (dynamics enforced), f (filtration), g (guidance), a (action trajectory post-optimization), and op (number of inner optimization steps).} \label{table:ablation}
}

\begin{tabularx}{0.47\textwidth}{ccccc|c|ccc}
\toprule
\multicolumn{5}{c}{\textbf{Features}} & \textbf{Loss} &  \multicolumn{3}{c}{\textbf{Metrics}}   \\ 
dyn & f & g & a & op & action only? & rule & real & fail  \\ 
\midrule
 $\surd$ & $\surd$ & $\surd$ & & 1 & & 0.0189 & 0.3588 & \textbf{0.1647} \\
 \midrule
  $\surd$ & & & & 1 & & 1.3802 & 0.3964 & 0.3011 \\
   $\surd$ & & $\surd$ & & 1 &  & 0.0387 &	0.4336 & 0.2227  \\
   $\surd$ & $\surd$ &  & & 1 &  & 1.3268 &	0.3869 & 0.2942 \\
    \midrule
  $\surd$ & & $\surd$ & $\surd$ & 1 &  & 0.019 &	0.447 &	0.1879 \\
  $\surd$ & $\surd$ & $\surd$ & $\surd$ & 1 &  & \textbf{0.0132}	& 0.4431 &	0.1845 \\
   $\surd$ & $\surd$ & & $\surd$ & 1 &  & 0.0329 & 0.4591 & 0.1881 \\
   $\surd$ &  &  & $\surd$ & 1 &  & 0.0819 & 0.4339	& 0.2117  \\
    \midrule
   $\surd$ &  $\surd$ & $\surd$ &  & 3 &  & 0.0158               & 0.4591               & 0.2017   \\
   $\surd$ &  $\surd$ & $\surd$ &  & 5 &  & 0.0156               & 0.456                & 0.1966  \\ 
 \midrule
 $\surd$  &  $\surd$ & $\surd$ &  & 1 & $\surd$ &  0.018 & \textbf{0.2057} & 0.2824 \\
  &  $\surd$ & $\surd$ &  & 1 &  & 0.0525	& 3.4204 &	0.612  \\
   &  &  &  & 1 &  & 1.2797               & 3.3251               & 0.7523  \\
\bottomrule
\end{tabularx}
\vspace{-5mm}
\end{table}

To analyze design choices, we conduct an ablation study under the speed limit rule setting. 
Results are shown in \Cref{table:ablation} where the top row is our proposed version of CTG.
The first section compares different combinations of guidance (g) and fitration (f).
Without guidance, rule violation increases greatly.
Filtration is more effective paired with guidance than by itself, as it can choose the best from several already-guided samples that may satisfy rules to differing degrees.
In the next part of the table, variants using an additional output action optimization (a) are evaluated.
Replacing guidance with optimization is worse on all metrics, while combining guidance with optimization reduces rule violation at the cost of higher realism deviation and failure rate.
Next, more inner optimization steps (op) are used in guidance, improving rule violation but giving worse realism and failure rate.
Finally, in the bottom section, we evaluate a variant that only supervises the action trajectory, and variants where unicycle dynamics (dyn) are not enforced.
Supervising only actions (instead of states and actions) gives more faithful accelerations resulting in lower realism deviation, but failure is more frequent without state supervision to regularize. Additionally, we find that enforcing dynamics using the unicycle model is key: all metrics degrade without this.

\section{Conclusion}
We proposed \method, a conditional diffusion model for the task of controllable traffic simulation, which opens several exciting future research directions.
Currently, we have only used \method to model vehicles, but cyclists and pedestrians are also important agents to simulate for AV interactions. Additionally, using collision and off-road guidance to enable very long-term, robust traffic simulation is an important application.
Outside of AV, the proposed guidance framework may benefit many tasks where learned models in-the-loop must be reactive and follow novel objectives online.

\noindent\textbf{Acknowledgments.}
The authors thank Or Litany, Sanja Fidler, and Karen Leung for valuable discussions and feedback.

\bibliographystyle{ieeetran}
\bibliography{root}

\addtolength{\textheight}{-6cm}   %

\end{document}